# A New Approach for Resource Scheduling with Deep Reinforcement Learning

Yufei Ye, Xiaoqin Ren, Jin Wang, Lingxiao Xu, Wenxia Guo, Wenqiang Huang and Wenhong Tian[*]
University of Electronic Science and Technology of China, Chengdu, China
tian_wenhong@uestc.edu.cn

## ABSTRACT
With the rapid development of deep learning, deep reinforcement learning (DRL) began to appear in the field of resource scheduling in recent years. Based on the previous research on DRL in the literature, we introduce online resource scheduling algorithm DeepRM2 and the offline resource scheduling algorithm DeepRM_Off. Compared with the state-of-the-art DRL algorithm DeepRM and heuristic algorithms, our proposed algorithms have faster convergence speed and better scheduling efficiency with regarding to average slowdown time, job completion time and rewards.

## CCS Concepts
• **Computing methodologies**➝**Heuristic function construction**
• **Computing methodologies**➝**Discrete space search**
• **Computing methodologies**➝**Computational control theory.**

## 1. INTRODUCTION
Resource management is a basic problem in the computer field. For example, resource management is particularly important in the field of cloud computing [1]. Due to many factors that need to be consider, the scheduling problem in most cases is an NP-hard problem or an NP-complete problem. The general solution step is to design an efficient heuristic algorithm under the specified conditions. Adjust the heuristic algorithms later in practice for better performance. However, due to the fixed nature of the heuristic algorithm, if certain restrictions change, it is usually necessary to repeat the above process.

Previous studies have shown that machine learning can provide a viable alternative to resource management heuristics [2]. The main method at the time is using reinforcement learning. With the rapid development of deep learning in the past few years, deep reinforcement learning has begun to emerge in various fields. For example, playing video games with humans [3], or playing chess with world champion [4]. Can deep reinforcement learning be applied to resource management? One successful case is the DeepRM [5]. DeepRM is set up in an environment where online jobs arrive one after another, and once the job is selected, it cannot be preempted. DeepRM can be used to learn to optimize a variety of different goals, such as minimizing average job slowdown or minimizing the job completion time. DeepRM effectively visualizes the state of resource occupancy and the status of waiting jobs. It is then possible to use deep learning methods to process its images and use reinforcement learning agents to make decisions and allocate resources to different jobs.

DeepRM performs simulation experiments on a synthetic data set. The results show that under a wide range of loads, DeepRM performance is even better than standard heuristics such as Shortest-Job-First (SJF) and a packing scheme inspired by Tetris [6].

The emergence of DeepRM is a good start for resource management with reinforcement learning, but it still has some room to improve. Based on DeepRM, this paper proposes some new improvements and ideas. First, using the behavior cloning in imitation learning [7] before starting deep reinforcement learning training. Then, changing the fully connected network to a convolutional neural network that facilitates image feature extraction. In addition, redefining the capacity of the cluster. By adopting the above method, the result of resource scheduling is improved. For example, the learning curve of Slowdown converges faster and decreases even more. At the same time, by modifying DeepRM's model design, it can be used not only for online resource scheduling but also for offline resource scheduling. These results will be conducive to the future research of computer network resource scheduling research.

## 2. BACKGROUND
We briefly review the aforementioned reinforcement learning and imitation learning.

### 2.1 Reinforcement Learning
Reinforcement learning [8] is an important branch of machine learning. Its essence is to solve the problem about making decision, that is, to automatically make decisions and make continuous decisions. It mainly consists of four elements, agent, environment, action, reward, and the goal of reinforcement learning is to obtain the cumulative rewards.

Due to learning through actual environment interaction, the model-free method is selected. The reinforcement learning model-free algorithm can be divided into a value-based method and a policy-based method according to the method for solving the optimal policy. The value-base method refers to solving the optimal policy by solving the optimal action value function. The policy-base method, like supervised learning, directly fits the policy.

### 2.2 Policy gradient methods
The policy gradient is a representation of the policy-based approach. The gradient of the objective given by [8]:

$$\nabla_\theta E_{\pi_\theta}\left[\sum_{t=0}^{\infty} \gamma^t r_t\right] = E_{\pi_\theta}[\nabla_\theta \log \pi_\theta(s,a) Q^{\pi_\theta}(s,a)] \qquad (1)$$

In formula (1), $Q^{\pi_\theta}(s,a)$ represents the cumulative reward value based on the $\pi_\theta$ policy. $\pi_\theta(s,a)$ represents the probability of selecting action *a* in state *s* under the current policy. After obtaining the gradient, we use the gradient ascent algorithm to update the parameters of the policy:

$$\theta \leftarrow \theta + \alpha \sum_t \nabla_\theta \log \pi_\theta(s_t, a_t) v_t \qquad (2)$$

Formula (2) is the process of parameter updating, $\theta$ is the policy parameter and $\alpha$ is the learning rate. $v_t$ is an unbiased estimate of $Q^{\pi_\theta}(s, a)$. Below, we specifically explain the entire steps.

## 2.3 Imitation learning

In traditional reinforcement learning tasks, the optimal policy is usually learned by calculating cumulative rewards. This method is simple and straightforward, and it has better performance when more training data is available. However, in a sequential decision, the learner cannot be rewarded frequently, and there is a huge search space based on cumulative rewards and learning methods. After years of development, the imitation learning method has been able to solve multi-step decision-making problems well, and speed up the training process of reinforcement learning greatly [9]. At present, imitation learning has been widely applied in robotics [10], Self-driving cars [11] and other fields.

The search space for multi-step decision-making in reinforcement learning tasks is enormous. It is very difficult to learn appropriate decisions before many steps based on cumulative rewards. Directly imitating the "state-action" of human experts can significantly ease this problem. We call this "Behavior Cloning".

Through the samples we provide, let the machine learn. Using a classifier or regression algorithm to learn the policy model, this policy can be used as an initial policy for machine reinforcement learning. Then through reinforcement learning methods based on environmental feedback to improve, so as to obtain a better policy.

## 3. DESIGN

In this section, we propose two kinds of design based on deep reinforcement learning, a kind of continuous online resource scheduling design of DeepRM, and the other is the design of offline resource scheduling modified based on DeepRM. At the same time, we have made corresponding changes to the structure of DeepRM's neural network. We replace the original simple BP neural network with a multi-layer convolutional neural network that is more conducive to image feature extraction.

## 3.1 Model

We consider a cluster with two resources, including CPU resources and memory resources. The arrival of the job can be divided into two situations: During online resource scheduling, the job will arrive at the waiting slot with a Poisson process one after another; when the offline resource is scheduled, the job reaches the waiting slot at one time. The resource requirements of the job are known on arrival. It includes the demand for CPU and memory, and the duration of the job. For simplicity, once the job is selected, it cannot be interrupted. At the same time, the clusters in our model are considered as a collection, ignoring the impact of fragmentation.

We have two different objective. One is to minimize the average job slowdown, and the other is to minimize the job completion time of the experiment. For the previous objective, we set the slowdown for each job to be given by $S_j = C_j / T_j$. In the formula, $T_j$ represents the ideal completion time of the job, $C_j$ represents the actual completion time (the sum of the ideal completion time and the job waiting time). Note that $S_j \geq 1$. Another objective is to minimize the completion time of the experiment, that is, to minimize the time from the time the first job reaches the job slot to the time the cluster completes all jobs.

## 3.2 Reinforcement learning formulation

### 3.2.1 State space

We use the image to show the status of the system every moment. The online scheduling system is slightly different from the offline scheduling system (see figure 1 for illustration).

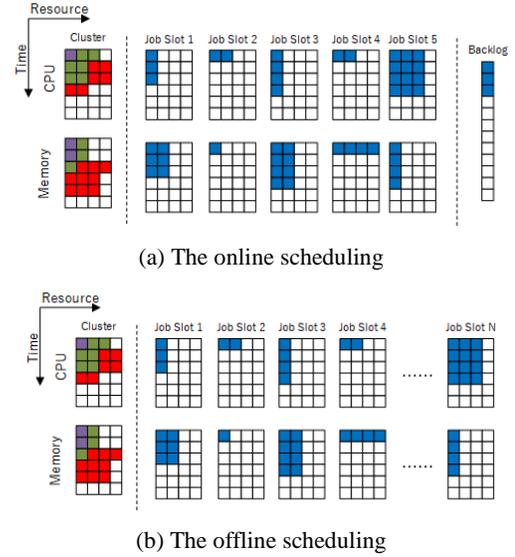

(a) The online scheduling

(b) The offline scheduling

**Figure 1. The state representation of online scheduling and offline scheduling**

Figure 1(a) shows online scheduling system consists of three parts: cluster, job slots and Backlog. The leftmost image with multiple color squares is the cluster, which represents CPU and memory resources. The squares in the cluster image indicate that they have been selected for scheduling. Different colors indicate different jobs. The horizontal axis of the image indicates the number of units of required resources, and the vertical axis indicates the next few time steps. The second part is the job slot. The role of job slots is to store jobs that are waiting to be scheduled. The horizontal length of the job represents the unit demand for the resource, and the vertical length represents the ideal completion time step. If the job slot is full of jobs, the remaining jobs will be placed in the backlog. The backlog only shows the number of jobs and does not display specific information.

Figure 1(b) shows the status of offline resource scheduling, which is different from online resource scheduling in two aspects. On the one hand, the number of job slots is far greater than the latter, because all jobs need to be loaded into job slots at the beginning of the schedule. On the other hand, since all jobs have been loaded, there are no more jobs. Since Backlog is not needed, we abandoned it.

### 3.2.2 Action space

At some point, the cluster may wish to schedule multiple jobs in the job slot. Whether it is online resource scheduling or offline resource scheduling, we set the maximum value of the action as the number of job slots plus one. When the action selects the last one, it means that no job is selected, otherwise the job corresponding to the action will be selected. When the selected job slot is empty, the cluster is full or the last action is selected, the system performs a "Move on" operation. "Move on" indicates that the selection of this moment is completed. The cluster executes a time step, that is, the cluster image moves upward one step. The "Move on" operation separates the action from the actual selection at the same moment so that the agent can maintain the linearity of the action space.

### 3.2.3 Rewards

Since we have two objective, we have two sets of rewards. When our objective is to minimize the average slowdown, our reward value for each time step is set to $\Sigma_{j\in\mathcal{J}}\frac{-1}{T_j}$, Where $\mathcal{J}$ is the current set of scheduled jobs and waiting jobs in the system. When our objective is to minimize the completion time of the experiment, our reward value for each time step changes to $-|\mathcal{J}|$, which is negative the number of unfinished jobs in the system.

## 3.3 Training process

Deep reinforcement learning is different from reinforcement learning because it uses neural networks to make decisions. DeepRM uses a hidden layer of fully connected neural networks as a policy. The above state space is the input of the network and outputs the probability distribution of all the actions. We do not change the input and output of the network, just change the way the network is connected. Because the input is a picture, we decided to try a convolutional neural network that is mostly used in image processing.

In order to speed up the training process of reinforcement learning, we prefer to use the imitation learning algorithm first. We decided to use the SJF algorithm as a mock object. Behavior cloning is similar to supervised learning. We think of multiple instances of job arrival sequences in the training process, hereinafter referred to as jobsets or experiments. We first randomly generated some jobs as jobset data, treating them as data samples in supervised learning. Then according to the algorithm of SJF, the corresponding scheduling order of the samples is calculated, and these sequences are considered as the labels of the samples. Finally, the results obtained are divided into two groups: training set and test set. Put the training set into our own defined convolutional neural network. The training is terminated and the model parameters are saved when the accuracy of the validation set stops increasing. Figure 2 shows the pseudo-code for the imitation learning. We maintain the trained neural network as an initial policy for deep reinforcement learning. When the initial policy is completed, we will use the policy gradient method mentioned above.

```
1:   for each iteration :
2:       π = SJF
3:       for each jobset :
4:           {s_1, a_1, s_2, a_2, s_3, a_3, ..., s_L, a_L} ~ π
5:       end
6:       Δθ ← 0
7:       for t = 1 to L :
8:           compute eval action : a_t^θ = π_θ(s_t)
9:       end
10:      Δθ ← Δθ − α∇_θ Σ_{i=1}^{t} a_t log a_t^θ
11:  end
```

**Figure 2. Pseudo-code for the imitation learning**

## 4. EVALUATION

We mainly observe the difference between different schemes from three aspects: the average job slowdown of the experiment (including the speed of convergence), the job completion time of the experiment, and the average and maximum value of the discount cumulative reward for the experiment during training.

## 4.1 Methodology

### 4.1.1 Workload

We use the workload setting as in [5]. We assume there are two types of resources in the cluster. The capacity represents CPU and memory in {$1r$, $1r$}. The resource requirements for each job are as follows: Each job randomly selects one resource as the primary resource, and the rest is the secondary resource. The demand for primary resources is between $0.5r$ and $1r$, and the demand for secondary resources is between $0.1r$ and $0.2r$. The duration of each job is as follows: 80% of the jobs are short-term jobs with a duration of $1t$-$3t$; 20% of the jobs are long-term jobs with a duration of $10t$-$15t$. During online resource scheduling, we can control the arrival rate of the job to change the load of the cluster capacity from 10% to 190% because the job is arrived according to the Poisson process.

### 4.1.2 DeepRM2

We use DeepRM as a prototype to set up a new generation online scheduling model named DeepRM2. In DeepRM2, We set the job slot to 10 units. The horizontal and vertical lengths of the cluster in Figure 1(a) are all 20 units, which is slightly larger than DeepRM. Each experiment lasts for 50 time steps. The number of job slots is set to 10, and the maximum number of backup logs is set to 60. During an Epoch period, we have 100 experiments. In each experiment, we run 20 Monte Carlo simulations. The biggest change in DeepRM2 is the use of a convolutional neural network structure, as shown in Table 1 below.

**Table 1. Convolutional neural network in DeepRM2**

| Layer name | Output size | DeepRM2 |
|---|---|---|
| Conv1 | 20x224 | 5x5, 8, stride 1, relu |
| Pool1 | 10x112 | 2x2 avg pool, stride 2 |
| Conv2 | 10x112 | 5x5, 16, stride 1, relu |
| Pool2 | 5x56 | 2x2 avg pool, stride 2 |
| FC1 | 1x1 | 72-d fc, tanh |
| FC2 | 1x1 | 11-d fc, softmax |

### 4.1.3 DeepRM_Off

We named offline resource scheduling algorithm as DeepRM_Off. We made a few changes in the system based on DeepRM2, including setting the number of job slots to the number of jobs per experiment so that each job can be loaded into the corresponding slot before scheduling. At the same time, we no longer need a backlog because the job just fits into the slot and there are no extra jobs.

## 4.2 Comparing online scheduling efficiency

Figure 3 shows the difference in average job slowdown between DeepRM2 and other methods under various loads. The experimental data is 100 random operations experiments that are not used for training. As can be seen from Figure 3: (1) Under all load conditions, there are three average work slowdowns that are always the lowest, followed by DeepRM2, DeepRM, and SJF in ascending order. It is learned that we know that the deep reinforcement learning algorithm can be better than the heuristic algorithm after training, and the optimized DeepRM2 is indeed superior to the original DeepRM. (2) Excluding the three methods previously compared, the slowdown results for Tetris and Random are both high. When the load is lower than 90%, Tetris is lower, whereas Random is lower. The reason is that Tetris will schedule large jobs so that small jobs cannot be scheduled in time to cause congestion.

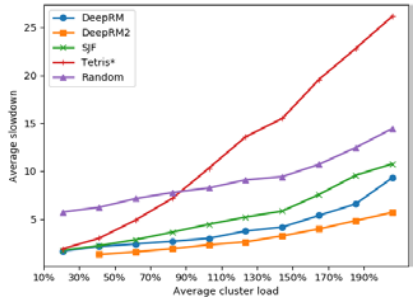

**Figure 3. Job slowdown under various loads**

Figure 4 shows how the corresponding result changes when we set different targets for DeepRM2 at the load of 70%. The corresponding target in the left and right parts of Figure 4 is the average job deceleration and job completion time of the experiment. Figure 4 shows that DeepRM2 is better than DeepRM in all aspects.

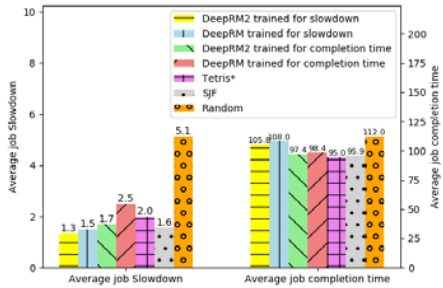

**Figure 4. Performance Comparison between DeepRM2 and other methods**

To demonstrate the advantages of DeepRM2, Figure 5 presents the discounted total reward value and the average job slowdown during the training of the original DeepRM and DeepRM2. Figure 5(a) and 5(b) show the situation where 500 epochs were trained under 100% cluster load. Only from the numerical comparison, in the final performance of DeepRM2, the discounted total reward defined in 3.2.3 is higher and the average slowdown is lower, so the overall performance is better. Under the same training condition, the training time of each epoch of DeepRM2 is 37.5% less than that of DeepRM, and DeepRM2 can train fewer rounds until convergence from the Figure 5. In general, through deep reinforcement training, both DeepRM2 and DeepRM are much better than the other three methods.

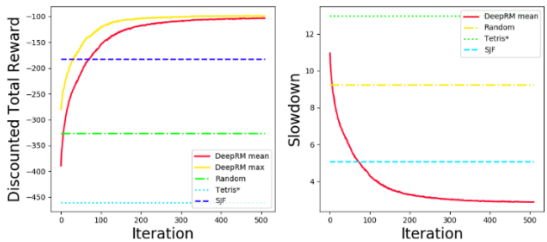

(a) DeepRM training process

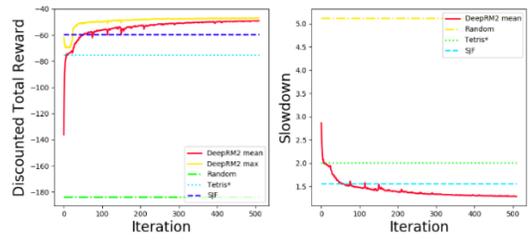

(b) DeepRM2 training process

**Figure 5. Discounted total reward value and the average job slowdown**

In order to explore how deep reinforcement learning reduces slowdown, we have made box plots for each job length before and after training for DeepRM2 and DeepRM. Figure 6(a) is the job slowdown after random scheduling, and Figure 6(b) is the job slowdown after training. Through different shades of color in Figure 6, it can be found that DeepRM2 is better than DeepRM under conditions compared. Then we specifically analyze the differences between (a) and (b).

When scheduling randomly, both short and long jobs have the same chance to wait because the cluster is full. However, due to the slowdown formula, it can be seen that when waiting for the same amount of time, the short jobs has a greater increase in slowdown than the long jobs. Therefore, the slowdown of the short job is large, and the slowdown of the long job is small. Since the short job has a higher proportion, the arithmetic average slowdown is high.

When the training is completed, the system learns that reducing the slowdown of the short job helps to reduce the slowdown of the arithmetic average job. However, DeepRM2 is better than SJF from the previous results. We speculate that DeepRM2 tends to schedule short jobs, but it has its own set of invisible standards. Not every short job will be scheduled firstly. Figure 6 (b) validates this view.

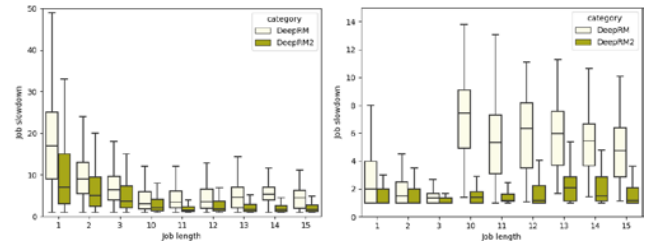

(a) The job slowdown after random scheduling     (b) The job slowdown after training

**Figure 6. The boxplots for each job length**

### 4.3 Comparing offline scheduling efficiency

The workload of offline scheduling is the same as online scheduling as mentioned in 4.1.1. The cluster and job slot configuration is described in 4.1.3. We briefly demonstrate the offline resource scheduling training process, as shown in Figure 7. Because DeepRM does not involve offline resource scheduling, DeepRM_Off are compared against the three traditional algorithms: SJF, Tetris, and Random. From the results, with the increase in the number of DeepRM_Off training, discount cumulative rewards have reached a maximum, the average job slowdown dropped to a minimum. Therefore, deep reinforcement learning can be applied to offline resource scheduling. This attempt is successful and effective.

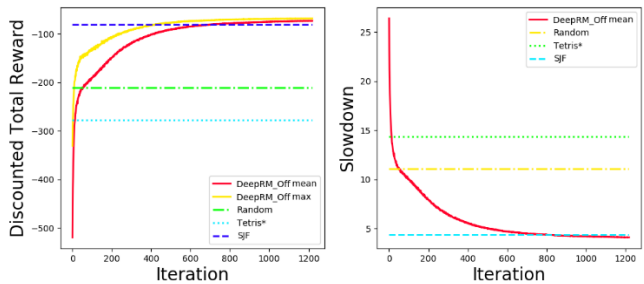

**Figure 7. Discounted total reward value and the average job slowdown of DeepRM_Off**

## 5. DISCUSSION

We currently did some work based on DeepRM, including modifying neural networks, expanding usage scenarios, and improving the convergence speed of training. However, many improvements can be further advanced. We will continue to try in the future so that the theoretical results can be applied to actual production too.

### 5.1 Multi-cluster

We are currently considering a single-cluster situation. In other words, we are putting patchwork node resources together and treating them as resources of a cluster. In reality, this usually does not work. Because jobs do not perfectly fill all the cluster space, each node will have some resource fragmentation that is difficult to use. We want to solve this problem in the future by considering a multi-cluster scheduling system.

### 5.2 Deep reinforcement learning algorithm

We mentioned in the previous section that we have improved the algorithm from full-connection to convolutional neural networks. But the algorithm we use is still Policy Gradient, we are ready to try more new algorithms, such as actor critic [12] algorithm that combines value-based and policy-based methods. The advantage of the Actor Critic method is that it can be updated step by step, faster than the traditional Policy Gradient. We can also choose the DDPG [13] algorithm, which not only combines the previously successful DQN structure [3], but also improves the stability and convergence of the Actor Critic.

## 6. RELATED WORK

Deep reinforcement learning has now been applied in various fields, including video games [3], Go [4], and resource scheduling [5]. Our work is mainly based on DeepRM [5]. They pioneered the use of advanced reinforcement learning applications in online resource scheduling. However, their model training time is not short. We decided to use the imitation learning [7] that has been applied in robotics [10] and Self-Driving [11] to increase training speed. In fact, it is indeed effective [9]. In addition, since deep reinforcement learning can be applied to online resource scheduling, we should also be able to apply it to offline resource scheduling for the same reason.

Weijia Chen et al. [14] also considered modifications to the deep learning network model, they also proposed the idea of using convolutional neural networks instead of full connections, but we think that the structure of the convolutional neural networks they proposed is very simple. We make a more complex design of the convolutional neural network structure so that it can better extract the state characteristics of reinforcement learning. Of course, besides the Policy Gradient [8] used, we will try more new reinforcement learning algorithms in the future, such as Actor-critic [12] or DDPG [13].

## 7. CONCLUSION

This article once again shows that it is feasible to apply the latest deep reinforcement learning technology to large-scale systems, including online scheduling and offline scheduling. The result of our proposed method DeepRM2 surpasses the state-of-the-art heuristic algorithms in some aspects. We are exploring some new methods to further improve the performance.